\documentclass[journal]{IEEEtran}

\usepackage{times}
\usepackage{graphicx}
\usepackage{subfigure}
\usepackage{parskip}
\usepackage{multirow}
\usepackage{amsmath}
\usepackage{color}
\usepackage[labelfont=bf,textfont=it]{caption}
\usepackage[lined,boxed,commentsnumbered, ruled]{algorithm2e}

\begin{document}
\title{Dual-Scale Single Image Dehazing Via Neural Augmentation}
%
%
%
\author{
Z. G. Li,~\IEEEmembership{Senior Member~IEEE,}, C. B. Zheng$^*$, H. Y. Shu, and S. Q. Wu,~\IEEEmembership{Senior Member~IEEE}
\thanks{* correspondence author}
\thanks{Zhengguo Li and Haiyan Shu are with SRO department, the Institute for Infocomm Research, Singapore, 138632, (emails: \{ezgli, hshu\}@i2r.a-star.edu.sg). Chaobing Zheng and Shiqian Wu are with the school of Information Science and Engineering, Wuhan University of Science and Technology, Wuhan 430081, China (e-mail: \{zhengchaobing, shiqian.wu\}@wust.edu.cn)}
}

%
%

\markboth{}
{Shell \MakeLowercase{\textit{et al.}}: Bare Demo of IEEEtran.cls
for Journals}
%



\maketitle

\begin{abstract}
Model-based single image dehazing algorithms restore haze-free images with sharp edges and rich details for real-world hazy images at the expense of low PSNR and SSIM values for synthetic hazy images. Data-driven ones  restore haze-free images with high PSNR and SSIM values for synthetic hazy images but with low contrast, and even some remaining haze for real-world hazy images. In this paper, a novel single image dehazing algorithm is introduced by combining model-based and data-driven approaches. Both transmission map and atmospheric light are first estimated by the model-based methods, and then refined by dual-scale generative adversarial networks (GANs) based  approaches. The resultant algorithm forms a neural augmentation which converges very fast  while the corresponding data-driven approach might not converge. Haze-free images are restored by using the estimated transmission map and atmospheric light as well as the Koschmieder's law. Experimental results indicate that the proposed algorithm can remove haze well from real-world and synthetic hazy images.
\end{abstract}

\section{Introduction}
Visual signals are distorted in adverse weather conditions which are usually classified as dynamic (such as rain and snow) or steady (such as haze, mist, and fog) \cite{1nara2003,1garg2004}. This paper focuses on the haze issue. Due to the effect of light scattering through small particles accumulated in the air, hazy images suffer from contrast loss of captured objects \cite{1nara2003},  color distortion \cite{1ancuti2013,He}, and reduction of dynamic range \cite{1zheng2013,1kouf2018}. Due to rain-streak accumulation in line of sight, haze is also an issue for heavily  rainy images  \cite{1garg2004}. Existing computer vision relevant such as object detection algorithms might not perform well on hazy images, especially for those real-world images with heavy haze or rain.  It is thus important to study single image dehazing.

Single image dehazing is widely studied because of its broad applications. Two popular types of single image dehazing algorithms are  model-based ones \cite{He,1fattal2008,1zhuq2015,CVPR16,1LiZG2021} and data-driven ones \cite{Ren,cai,ICCV17,1qin2020,1dongh2020,Qu_2019_CVPR}. The model-based ones are on top of the Koschmieder's law \cite{koschmider}. They can improve the visibility of real-world hazy images well regardless of haze degree but they usually cannot achieve high PSNR and SSIM values on synthetic sets of hazy images. On the other hand, the data-driven ones perform well on the synthetic sets while their performance could be poor for real-world hazy images, especially for those images with heavy haze \cite{glots2020,1liuw2021}. It is desired to have a single image dehazing algorithm which is applicable for both synthetic and real-world hazy images without sacrificing visual quality.

In this paper, a novel neural augmented single image dehazing algorithm is proposed through integrating the model-based and data-driven approaches.    Same as the model-based methods in \cite{He,1fattal2008,1zhuq2015,CVPR16,1LiZG2021} and the data-driven ones such as \cite{cai},  the atmospheric light and transmission map are required by the proposed algorithm to restore the haze-free image. Rather than using the model-based methods in \cite{He,1zhuq2015,CVPR16,1LiZG2021} or the data-driven methods in \cite{cai,1chen2021}, both the atmospheric light and the transmission map are first estimated by using model-based methods, and then refined by data-driven approaches. The atmospheric light  is  estimated  by using the hierarchical searching method in \cite{Three} which is derived from the Koschmieder's law \cite{koschmider}.  The transmission map is initialized by using the dark direct attenuation prior (DDAP) in \cite{1LiZG2021}, and is subsequently processed by the haze line averaging (HLA) algorithm in \cite{1LiZG2021} to alleviate morphological artifacts caused by the DDAP.

The atmospheric light and transmission map are then refined via data-driven approaches. Pre-trained deep neural networks (DNNs) can be served as fast solvers for complex optimization problems \cite{1hornik1989,1nir2021}. Such an advantage of data-driven approaches is well utilized by our refinement.  Since it is very difficult or even impossible to have a pair of a real-world hazy image and its corresponding haze-free image, the popular generative adversarial network (GAN) \cite{1gan2014} is utilized to refine the atmospheric light and transmission map. Both the generator and the discriminator are dual-scales and are based on the  Laplacian pyramid hazy image model  in \cite{1LiZG2021}. They are different from the single-scale generator and discriminator in \cite{1Lizg2022}. The atmospheric light and transmission map estimated by the model-based methods can be regarded as noisy atmospheric light and transmission map. Therefore, the main function of the proposed dual-scale refinement is to reduce/remove the noise from the atmospheric light and transmission map estimated by the model-based methods.  Based on this observation, the generator of the proposed GAN is constructed from the recursive residual group (RRG) in \cite{CycleISP}. Deep neural networks (DNNs) are usually biased towards learning low-frequency functions \cite{1raha2018}. Shortcut connections are thus adopted in the proposed GAN to preserve the high-frequency information in the dehazed image. The discriminator of the proposed GAN is based on the PatchGAN in \cite{1isola2017}.  To reduce the training cost, the proposed dual-scale GAN is trained by only using 500 hazy images which are generated from 500 realistic images in the realistic single image dehazing (RESIDE) datasets in \cite{1LiB2018}. Depth is estimated by using the algorithm in \cite{1lizq2018}. The hazy images are synthesized by using the Koschmieder's law \cite{koschmider}. This is different from the training of the model-based deep dahazing algorithm in \cite{1Lizg2022}. Besides the 500 synthesized hazy images,  500  hazy images from the multiple real-world foggy image defogging (MRFID) dataset  in  \cite{1liuw2021} are also utilized to train the algorithm in \cite{1Lizg2022}. Existing data-driven dehazing algorithms such as \cite{1qin2020,1dongh2020,1zhao2021,1dong2020} are trained by using more than 10K hazy images.

Four loss functions are applied to train the proposed dual-scale GAN. Two single-scale loss functions are defined by using an extreme channel \cite{1Lizg2022} and  the gradients of the restored and ground-truth images, respectively.  Besides them, the dual-scale adversarial loss function \cite{1gan2014} and dual-scale $l_1$ loss function are also utilized to train the proposed GAN. It is worth noting that the adversarial loss function  usually produces sharper images but with lower PSNR and SSIM values than the $l_1$ and $l_2$ loss functions. Since the haze-free image is restored by the Koschmieder's law, the high visual quality property from model-based approaches is well preserved. The model-based estimation and the data-driven refinement form a neural augmentation \cite{1nir2021,2zheng2020,1zheng2020}.  Theoretical analysis in the appendix indicates that the proposed neural augmentation framework converges faster than the corresponding data-driven approach. Experimental results show that the proposed algorithm is applicable to  synthetic and real-world hazy images, and outperforms existing data-driven dehazing algorithms for the real-world hazy images from the dehazing quality index (DHQI) \cite{1min2019} and fog aware density evaluator (FADE) \cite{1choi2015} points of view. Overall, four contributions of this paper are: 1) a novel neural augmentation which combines model-based and data-driven approaches for single image dehazing.  The framework is applicable to synthetic and real-world hazy images. The number of training data  can also be reduced significantly;  2) a new initiative on analyzing the neural augmentation theoretically by leveraging control theory \cite{1lizg2001,1khalil2002}. The analysis indicates that the neural augmentation framework converges faster than the corresponding data-driven approach if the model-based method is accurate. 3) generator and discriminator on top of a dual-scale dehazing algorithm which can preserve the high-frequency information of the restored haze-free image better than that on top of a single-scale dehazing algorithm; and 4) new loss functions for the deep-learning based single image dehazing.

The remainder of this paper is organized as below. Relevant works on single image dehazing are summarized  in Section \ref{relatedworks}. Details of the proposed algorithm are presented in Section \ref{newalgorithm}.  Experimental results are provided  to verify the proposed algorithm in Section \ref{experiment}.  Finally, conclusion remarks are  provided  in Section \ref{conclusion}.

\section{Related Works}
\label{relatedworks}

Let $L_c(p,s)$ be the light intensity of pixel $p(=(i,j))$ at distance $s$ for a color channel $c\in\{R, G, B\}$, $\tau(s)$ be the extinction coefficient which defines the rate that light is occluded, and  $A_c$ be the atmospheric light. By integrating the following particle model
\begin{align}
\frac{\partial L_c(p,s)}{\partial s}=\tau(s)(A_c-L_c(p,s)),
\end{align}
it can be derived that
\begin{align}
\label{initialmodel}
L_c(p, d(p)) = L_c(p,0)t(p)+A_c(1-t(p)).
\end{align}
Here $d(p)$ is the depth of pixel $p$. $t(p)$ is the transmission map, and it is computed as
\begin{align}
t(p)=\exp^{-\int_{0}^{d(p)}\tau(s) ds}.
\end{align}
For simplicity, let $L_c(p, d(p))$ and $L_c(p,0)$ be denoted as $Z_c(p)$ and $I_c(p)$, respectively.  They are a hazy image and the corresponding haze-free image. The model (\ref{initialmodel}) becomes the following Koschmieder's law \cite{koschmider}:
\begin{equation}
\label{eq1}
Z_c(p) = I_c(p)t(p) + A_c(1-t(p)).
\end{equation}
When the atmosphere is  homogenous, $t(p)$ can be computed by $\exp^{-\alpha d(p)}$, in which $\alpha(>0)$ represents the scattering coefficient of the atmosphere. The term $I_c(p)t_c(p)$ is called direct attenuation, and the term $A_c(1-t(p))$ is called airlight.
By using the equation (\ref{eq1}), both visibility and color saturation of a hazy image are reduced. On the other hand, under-exposed and over-exposed regions of the image $I$ could become well-exposed ones in the hazy image $Z$  due to the airlight, and the dynamic range of $Z$ is reduced \cite{1zheng2013,1kouf2018}. Generally, the visual quality of a hazy image becomes poor if the haze is heavy \cite{Li2015}. It is thus important to restore the haze-free image $I$.

Single image dehazing is a challenging problem because the transmission map $t$ depends on the unknown and varying depth. There are two popular types of algorithms:  model-based and data-driven ones. Many model-based dehazing algorithms were proposed by using the model (\ref{eq1}). Both the atmospheric light $A$ and the transmission map $t$ need to be estimated by the model-based methods. Existing methods on estimation of the atmospheric light such as \cite{He,Three} are derived from the model (\ref{eq1}). The $A$ is assumed to be fixed in the model-based methods while it will be refined by a data-driven approach in the proposed neural augmentation. Since the number of freedoms is larger than the number of observations, single image dehazing is an ill-posed problem.  Different priors were proposed to reduce the number of freedoms \cite{He,1fattal2008,1zhuq2015,CVPR16}. Among them, the dark channel prior (DCP) in \cite{He} is the most popular one even though it  introduces morphological artifacts to the restored image. Guided image filter (GIF) \cite{1he2013} and  weighted GIF (WGIF) \cite{Li2014} are good candidates to reduce the morphological artifacts due to their simplicity. A color-line prior was proposed in \cite{1fattal2008} by using an observation that small image patches typically exhibit a 1D distribution in the RGB color space. The haze line prior (HLP) in \cite{CVPR16} is based on an observation that a finite number of different colors  which are classified into clusters can represent colors of a haze-free image in the RGB space \cite{1orchard1991}. All the corresponding pixels in each cluster form a haze line in the hazy image. The HLP assumes that there exists at least one haze-free pixel in each haze line. The artifacts of the transmission map caused by the HLP are reduced by using the weighted least squares (WLS) framework \cite{1farb2008}.

\begin{figure*}[htb]
	\centering
	\includegraphics[width=0.86\textwidth]{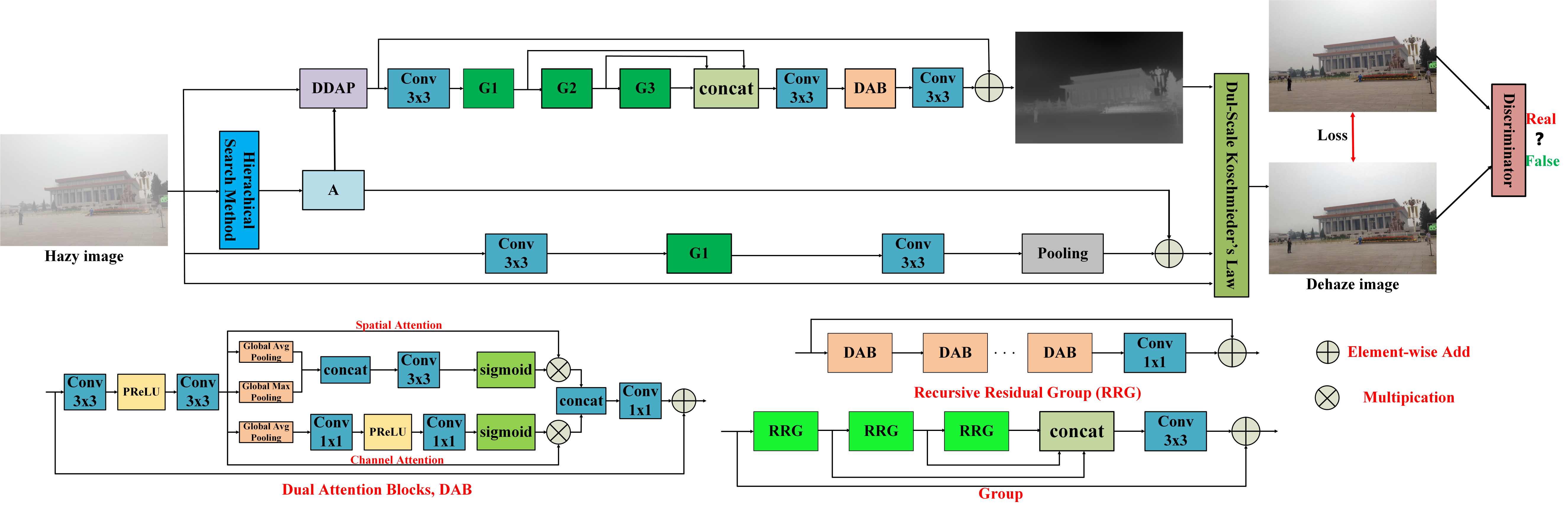}
	\caption{The proposed framework for dual-scale single image dehazing via model-based GAN. The generator is on top of the DNN in \cite{CycleISP}, and the discriminator is based on the PatchGAN in \cite{1isola2017}. The generator and discriminator are with two scales. Each dual attention block (DAB) \cite{CycleISP} contains spatial attention and channel attention modules. The Recursive Residual Group (RRG) contains several DABs and one $3*3$ convolution layers.}
	\label{Fig3}
\end{figure*}

Although the  priors in \cite{He,CVPR16} are robust to  haze degree in hazy images, they are not true for those pixels in the sky region. The DDAP was extended from the DCP such that the prior is applied for all the pixels in a hazy image \cite{1LiZG2021}. In addition, these physical priors are not always reliable, leading to inaccurate transmission estimates and producing artifacts in the restored images. The concept of haze line in \cite{CVPR16} was adopted in \cite{1LiZG2021} to form the HLA algorithm which can reduce the morphological artifacts significantly. The transmission map is usually refined by solving a complex optimization problem \cite{He,CVPR16}. Fortunately, the optimal solution can be approximated by a pre-trained DNN \cite{1hornik1989,1nir2021}. The refinement of the transmission map in the proposed neural augmentation is on top of such an observation. Amplification of noise in the sky region could also be an issue for the model-based dehazing algorithms \cite{glots2020,Li2015}. Furthermore, PSNR and SSIM values of haze-free images restored by model-based dehazing algorithms are usually low for synthetic hazy images even though the restored images are with sharp edges and rich details for real-world hazy images.

There are many data-driven dehazing algorithms, especially deep learning based dehazing algorithms.  Convolutional neural networks (CNNs) based algorithms such as \cite{Ren,cai,1qin2020,1dongh2020} have potential to obtain high SSIM or PSNR values for synthetic hazy images but the dehazed images look blurry and they are not photo realistic.  Qu et al. \cite{Qu_2019_CVPR} proposed an interesting GAN based dehazing algorithm on top of the physical scattering model  \cite{koschmider}. Dong et al. \cite{1dong2020} proposed a fully end-to-end GAN with fusion discriminator (FD-GAN) which takes frequency information as additional priors for single image dehazing. The GAN based dehazing algorithm is able to produce photo-realistic images even though generated texture and real one could be different. This problem could be addressed by using the hazy image dataset in \cite{1liuw2021} to train the GANs. Large pairs of hazy and clean images are necessary for the training of the data-driven methods. Most data-driven methods are trained on synthetic hazy images \cite{1zhang2018,1shao2020}. Due to the domain gap between synthetic and real-world data,  recent investigations indicate that the data-driven algorithms perform well for synthesized hazy images but poor for real-world hazy images \cite{glots2020,1liuw2021}. This is because it is almost impossible to obtain the ground-true depth information in most real-world scenarios. It is worth noting that the DCP and GANs were used \cite{1yang2018,1li2020,1zhao2021} to address the problem of lacking paired training images for real-world hazy images. The morphological artifacts caused by the DCP could be an issue for these algorithms.

Model-based and data-driven approaches were originally fused together in \cite{2zheng2020,1zheng2020} to  form innovative neural augmentation frameworks for high dynamic range imaging. The idea was borrowed to develop neural augmented single image dehazing frameworks in \cite{1Lizg2022,1zhao2021}. The DCP \cite{He} was also applied to estimate the transmission map and atmospheric light in the RefineDNet \cite{1zhao2021}. Only the transmission map is refined by a data-driven approach in \cite{1zhao2021} while both the atmospheric light  and the transmission map are refined by using data-driven approaches in the proposed algorithm.  The GIF \cite{1he2013} was used in \cite{1zhao2021} to reduce the morphological artifacts caused by the DCP while the simple HLA algorithm in \cite{1LiZG2021} is used in the proposed algorithm. As pointed out in \cite{2zheng2020,1zheng2020}, the simplicity of the model-based method is crucial for the neural augmentation. A  perceptual fusion strategy is also adopted in the RefineDNet to blend two different dehazing outputs which further increases the complexity of the RefineDNet. In addition, the RefineDNet is trained by using 13,990 synthetic hazy images in the RESIDE datasets \cite{1LiB2018} while the proposed framework is trained by only using 500 hazy images. The neural augmentation framework in this paper extends the data-driven algorithm in \cite{1Lizg2022} from single-scale to dual-scale. It should be mention that the  neural augmentation framework was also introduced to study high dynamic range imaging in \cite{2zheng2020,1zheng2020}. However, there is no theoretical analysis on the neural augmentation framework in \cite{1Lizg2022,1zhao2021,2zheng2020,1zheng2020}. This paper also  provides new theoretical analysis on the framework in the appendix. The theoretical result actually includes an important guideline for the neural augmentation framework, i.e., the model-based method is required to be {\it as accurate as possible}.

\section{Neural Augmented Single Image Dehazing}
\label{newalgorithm}
Both the atmospheric light $A$ and the transmission map $t$ are  required by the proposed algorithm to restore the haze-free image $I$. They are first estimated by model-based methods, and then refined by data-driven approaches.  The model-based and data-driven approaches form a neural augmentation \cite{1nir2021,1Lizg2022,2zheng2020,1zheng2020}. The overall framework is shown in Fig. \ref{Fig3}. Mathematically, the atmospheric light and transmission map are computed as
\begin{align}
\label{softmax}
\left\{\begin{array}{l}
A_{\tau}= A_m+A_{d,\tau}\\
t_{\tau}(p)=t_m(p)+t_{d,\tau}(p)
\end{array}
\right.,
\end{align}
where $A_m$ and $t_m(p)$ are obtained by the model-based method.   $A_{d,\tau}$ and $t_{d,\tau}$ are the outputs of the data-driven approaches  for the atmospheric light and the transmission map, respectively. $\tau$ is the iteration of the proposed neural augmentation framework. For brevity of notation, the item $\tau$ is omitted in the main body of the paper.

The proposed framework (\ref{softmax}) can be explained by leveraging conventional knowledge from the field of nonlinear systems \cite{1lizg2001,1khalil2002} where modelled dynamics and unmodelled dynamics are two important concepts. Due to limited representation capability of the model-based approaches, the unmodelled dynamics  can be further represented by applying the data-driven approach \cite{1lizg2001,1khalil2002}.   Since the data-driven approaches only produce the correction terms $A_{d,\tau}$ and $t_{d,\tau}$ which are interleaved with the model-based terms $A_m$ and $t_m(p)$, the amount of training data required by the neural augmentation to achieve a given accuracy is notably smaller than that required by the corresponding data-driven approach \cite{1nir2021}.  As shown in the appendix, the proposed neural augmentation framework (\ref{softmax}) also converges faster than the corresponding data-driven approach if the model-based terms $A_m$ and $t_m(p)$ are accurate.

\begin{figure*}[!htb]
\centering{
\subfigure[a hazy image]{\includegraphics[width=1.3in]{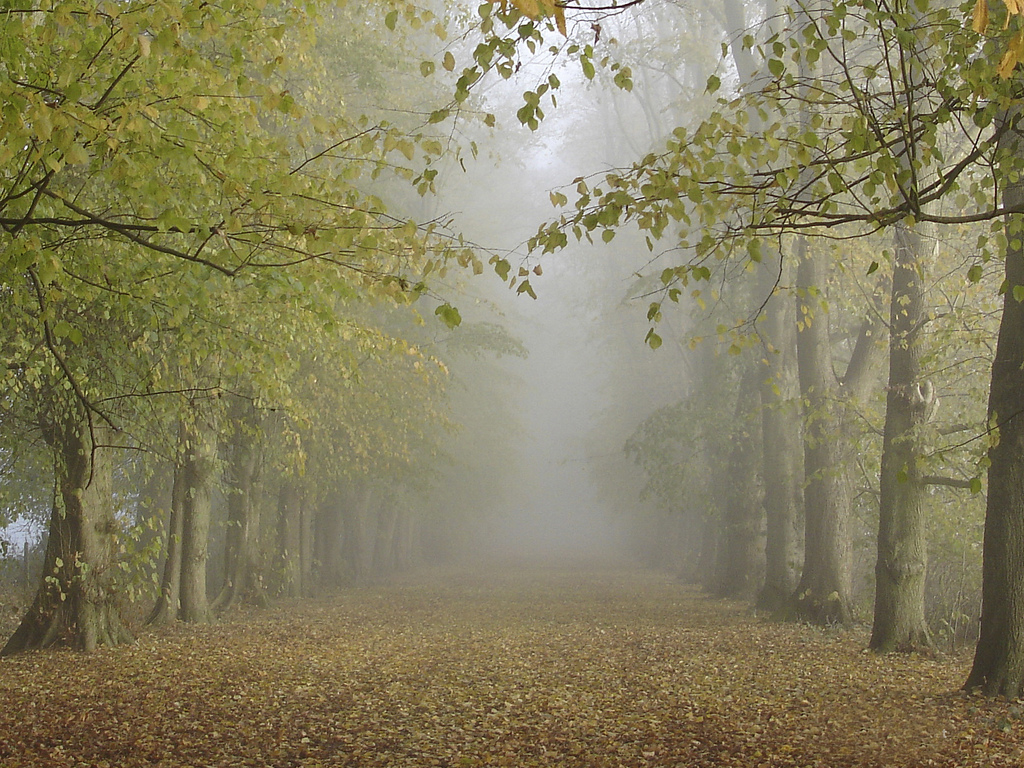}}
\subfigure[an initial transmission map $t_0$ ]{\includegraphics[width=1.3in]{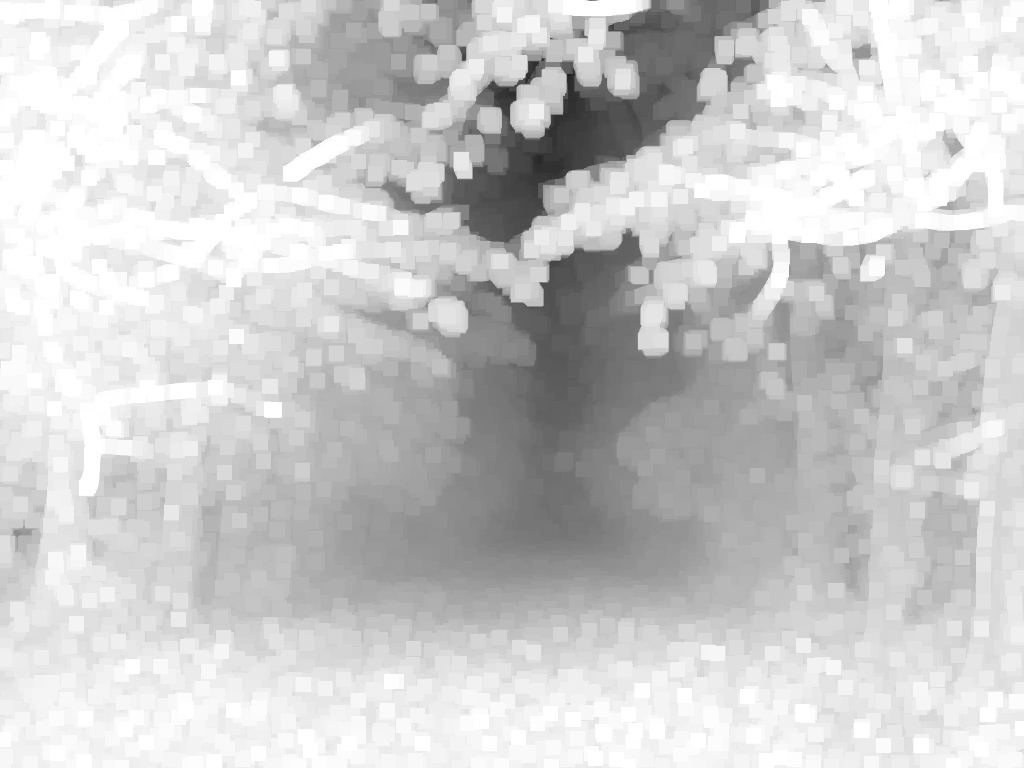}}
\subfigure[a refined transmission map $t_m$]{\includegraphics[width=1.3in]{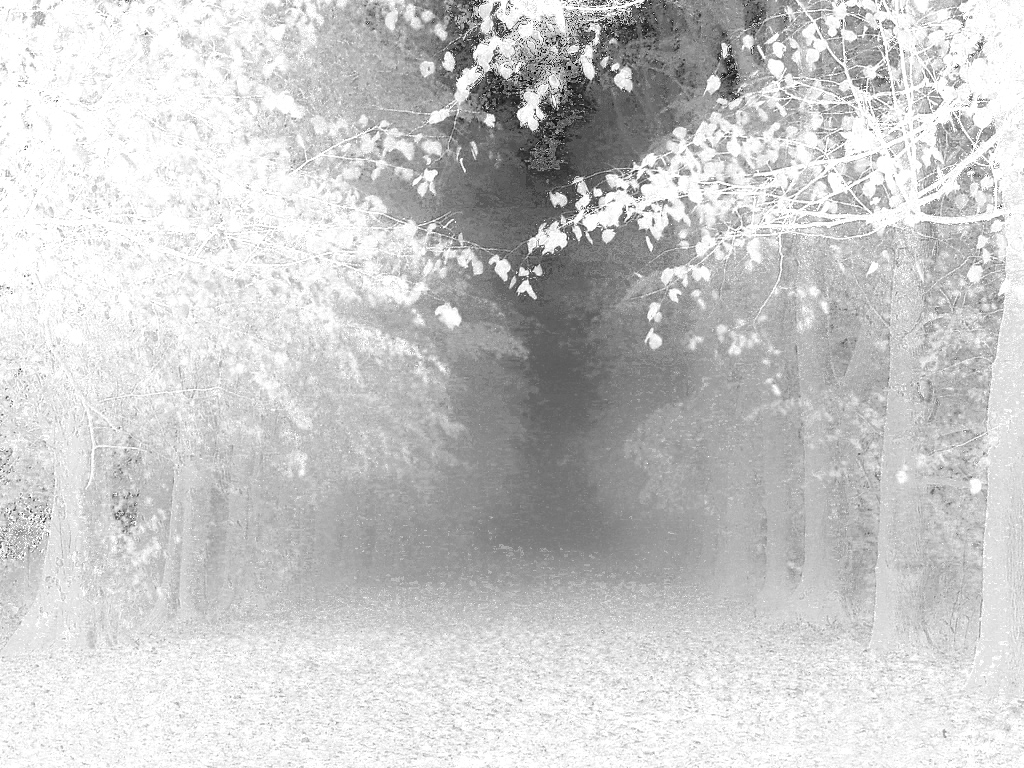}}
\subfigure[a dehazed image by the initial transmission map $t_0$]{\includegraphics[width=1.3in]{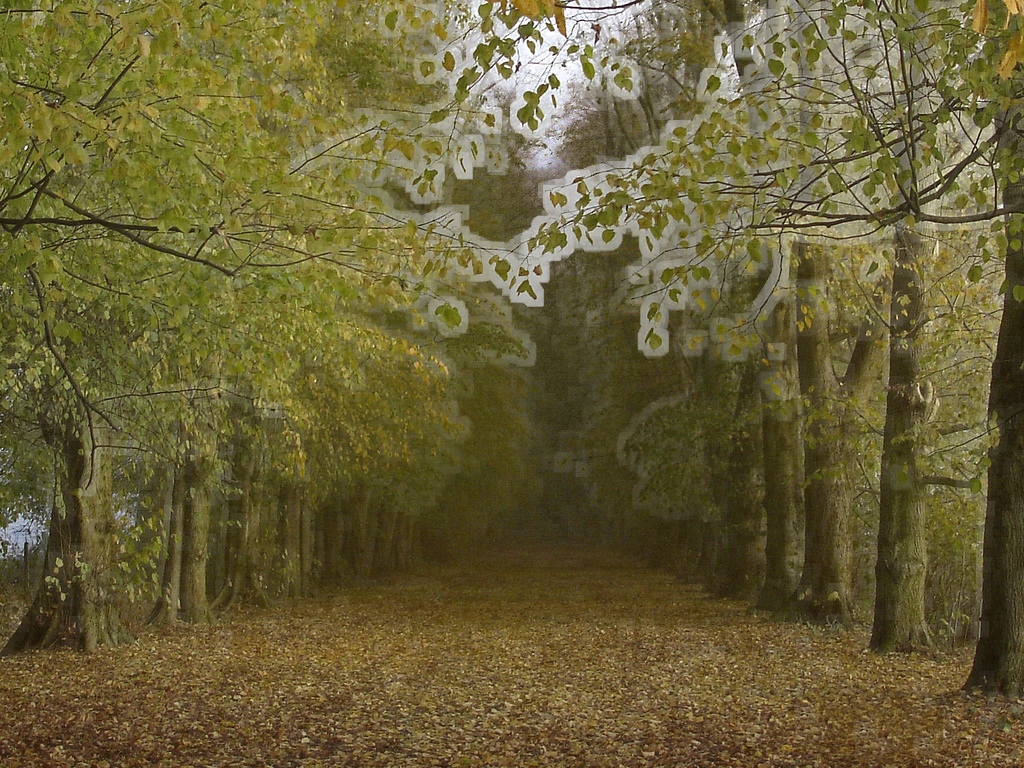}}
\subfigure[a dehazed image by the refined transmission map $t_m$]{\includegraphics[width=1.3in]{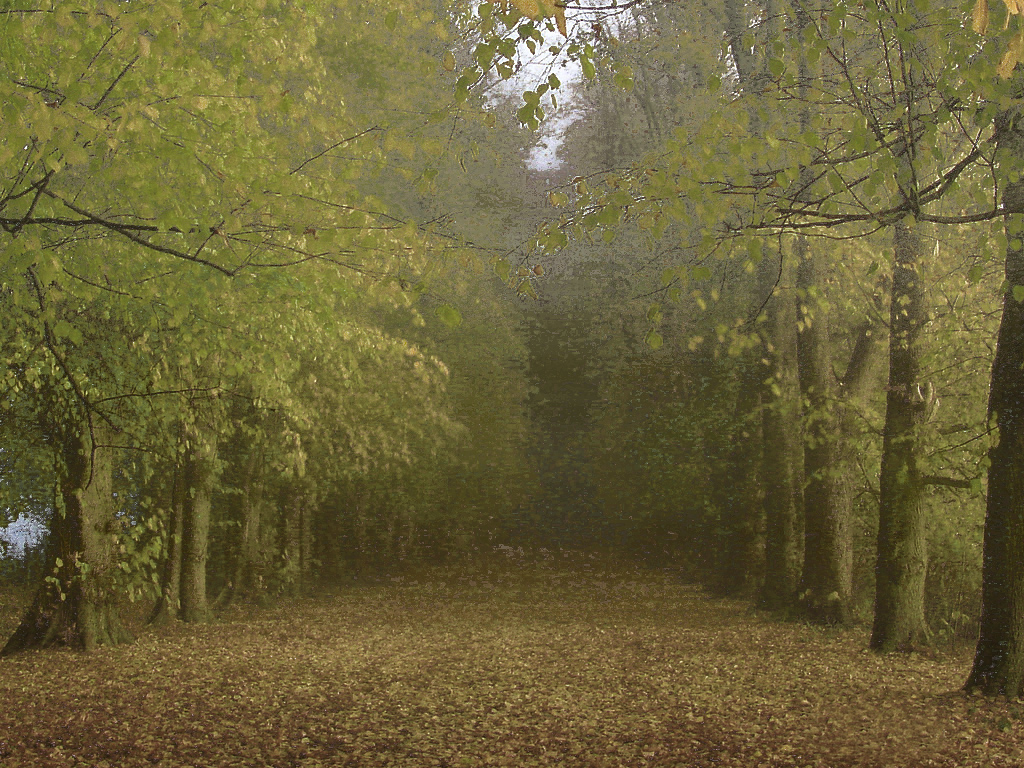}}}
\caption{Comparison of initial and refined transmission maps as well as their corresponding dehazed images.}
\label{FigFig2}
\end{figure*}

\subsection{Model-Based Estimation of $A$ and  $t$}
\label{initializationsubsection}

The model-based estimation of $A$ and  $t$ is on top of the method in \cite{1LiZG2021}. Details are provided such that this paper is self-contained.

\subsubsection{Estimation of $A$}

The hierarchical searching method in  \cite{Three} is adopted by the proposed framework to estimate an initial value of the atmospheric light $A$.
The hazy image $Z$ is divided into four rectangular regions. Let $\mu_{i,c}$ and $\sigma_{i,c}$ be  the average pixel value and the standard deviation of the pixel values for the color channel $c$ in the $i$th  rectangular region. The score of the $i$th region is defined as
\begin{align}
S_i=\frac{1}{3}\sum_{c\in \{R, G,B\}}(\mu_{i,c}-\sigma_{i,c}).
\end{align}
Let $i_0$ be $\arg\max_i\{S_i\}$. The $i_0$th region is further divided into four smaller regions. This process is repeated until the size of the selected region is smaller than a pre-specified threshold such as $32\times 32$. Within the finally selected region, the color vector $[Z_r(p^*), Z_g(p^*), Z_b(p^*)]$ is selected as the model-based atmospheric light $A_m$. Here, $p^*$ is obtained by
\begin{align}
\arg\min_{p\in S_{i_0}} \|[Z_r(p)-255, Z_g(p)-255, Z_b(p)-255]\|.
\end{align}

\subsubsection{Estimation of $t$}

The dark direct attenuation of a hazy image $Z$ is defined as  \cite{1LiZG2021}
\begin{align}
\label{rho}
\psi_{\rho}(I*t)(p)=\min_{p'\in \Omega_{\rho}(p)}\{\min_{c\in\{R, G, B\}}\{I_c(p')t(p')\}\},
\end{align}
where $\Omega_{\rho}(p)$ is a square window centered at the pixel $p$ of a radius $\rho$ which is usually selected as 7. By assuming $\psi_{\rho}(I*t)(p)$ as zero, an initial transmission map is computed  as
\begin{align}
\label{transmap}
t_0(p)=1- \psi_{\rho}(\frac{Z}{A})(p).
\end{align}

As shown in Fig. \ref{FigFig2}(d), there are visibly morphological artifacts if the $t_0(p)$ is directly applied to restore the haze-free image $I$. The HLA algorithm in \cite{1LiZG2021} is adopted to reduce the morphological artifacts. The HLA algorithm is on top of the following haze line  \cite{CVPR16}:
\begin{align}
\label{hijij}
H(p)=\{p'|\sum_{c\in \{R, G,B\}}|I_c(p')-I_c(p)|=0\}.
\end{align}

 The haze line $H(p)$ can be identified by using color shift hazy pixels  $(Z(p')-A)$'s \cite{CVPR16}. The pixel $I_c(p)$ and the atmospheric light A are the two end points of the haze line $H(p)$. Let $\|Z(p')-A\|$ be denoted as $r(p')$. $1/r(p)$ is estimated as
\begin{align}
\frac{1}{r(p)}=\frac{1}{\mu_{sum}}\sum_{p'\in H_s(p)}\mu(p') \frac{t_0(p')}{r(p')},
\end{align}
where $\mu(p')$ is a weight, and $\mu_{sum}$ is $\sum_{p'\in H_s(p)}\mu(p')$. $H_s(p)$ is a subset of $H(p)$.

Choosing $\mu(p')$  as $r(p')$, the model-based estimation of $t(p)$ is given as \cite{1LiZG2021}:
\begin{align}
\label{myrrmaxij11}
t_m(p)=\frac{ \sum_{p'\in H_s(p)}t_0(p')}{\sum_{p'\in H_s(p)}r(p')}r(p).
\end{align}

As illustrated in Fig. \ref{FigFig2}(e),  the morphological artifacts are significantly reduced by the HLA algorithm (\ref{myrrmaxij11}).

\subsection{Data-Driven Refinement of $A$ and $t$}
\subsubsection{Structure of Data-Driven Refinement}
The GAN \cite{1gan2014} is utilized to refine the atmospheric light and transmission map as shown in Fig. \ref{Fig3}. Since it is much more challenging to refine the transmission map, the generator of the atmospheric light is simpler than the generator of the transmission map.

 As indicated in the introduction, the model-based terms $t_m(p)$ and $A_m$ are noisy. The generator is thus built up on top of the RRG module for the noise reduction in \cite{CycleISP}. Each RRG contains multiple dual attention blocks (DABs) which is composed of a spatial attention block and  a channel attention block. The DAB can suppress the less useful features and only allow the propagation of more informative ones. Therefore, it effectively deals with the uneven distribution of haze.  As indicated in \cite{1raha2018}, the DNNs are usually biased towards learning low-frequency functions. Therefore, shortcut connections are adopted  to connect RRGs in each group and connect groups in the proposed GAN.  The shortcut connections at both the RRG and group levels can retain shallow layers information and pass it into deep layers. Subsequently, the high-frequency information could be well preserved.  The discriminator is based on the PatchGAN in \cite{1isola2017}.

\subsubsection{Dual-Scale Single Image Dehazing}
A Laplacian pyramid is usually generated on top of a Gaussian pyramid  \cite{1burt1983}.
 Let $\{Z\}_G^l$ and  $\{Z\}_L^l$ be the $l(\in \{0, 1\})$-th level in the Gaussian and Laplacian pyramids of the image $Z$, respectively \cite{1burt1983}.   The Gaussian pyramid of  the transmission map $t$ is denoted as $\{t\}_G^l$. It can be easily derived that \cite{1LiZG2021}
  \begin{align}
 \label{newmodel}
 \left\{\begin{array}{l}
  \{Z\}_G^{1}(p)=(\{I\}_G^{1}(p)-A)\{t\}_G^{1}(p)+A\\
  \{Z\}_L^{0}(p)=\{I\}_L^{0}(p)\{t\}_G^{0}(p)
  \end{array}
  \right..
 \end{align}

The Laplacian pyramid of the haze-free image is restored via the following two-scale dehazing algorithm:
\begin{align}
\label{Iij}
 \left\{\begin{array}{l}
 \{I\}_G^{1}(p)=\frac{\{Z\}_G^{1}(p)-\varphi(A,\frac{3}{8})}{\varphi(\{t\}_G^{1}(p)),\frac{1}{4})}+\varphi(A,\frac{3}{8})\\
  \{I\}_L^{0}(p)=\frac{\{Z\}_L^{0}(p)}{\varphi(\{t\}_G^{0}(p), \frac{1}{4})}
  \end{array}
  \right.,
 \end{align}
where the function $\varphi(z,m)$ is defined as \cite{1wang2020}
\begin{align}
\varphi(z, m)=\left\{\begin{array}{ll}
|z|; &\mbox{if~}|z|\geq m\\
\frac{z^2+m^2}{2m}; &\mbox{otherwise}
\end{array}
\right.,
\end{align}
 and the function  $\varphi(z,m)$ is differentiable with respect to the variable $z$ \cite{1wang2020}.

The Laplacian pyramid $\{I\}_G^{1}$ and $\{I\}_L^{0}$ is collapsed to produce the haze-free image $I$. It can be easily verified that
\begin{align}
\varphi(\{t\}_G^{l}(p)),\frac{1}{4})\geq \frac{1}{8},\; ;\; l\in\{0,1\}.
\end{align}
Thus, the noise in the sky regions can be avoided from being amplified by using the dual-scale haze removal algorithm (\ref{Iij}) \cite{1Lizg2022}. The high-frequency information is also preserved better in the haze-free image $I$ by using the dual-scale structure.

\subsubsection{Loss Functions of Data-Driven Refinement}
Besides the structure of the proposed GAN, loss functions also play an important role in the proposed GAN. Since both the atmospheric light $A$ and transmission map $t(p)$ are refined by the GAN, the loss functions are defined by using the  restored image $I$ and the image $\{I\}_G^{1}$ as well as the
ground-truth image $T$ and the image $\{T\}_G^{1}$ in dual-scale.

 The first loss function is defined by using an extreme channel \cite{1Lizg2022}.  Let $\bar{I}_c(p)$ be defined as
\begin{align}
\bar{I}_c(p)=\min\{I_c(p),255-I_c(p)\},
\end{align}
and the extreme channel of the image $I$ is defined as
\begin{align}
\label{extremechannel}
\phi_{\rho}(I)(p)=\psi_{\rho}(\bar{I})(p).
\end{align}
The limitation of DCP on both the sky regions and high brightness objects can be avoided via the extreme channel.

Considering a pair of hazy image $Z$  and clean image $T$ which are captured from the same scene, the corresponding haze-free image of the hazy image is $I$. It can be known from the conventional imaging model \cite{1gonz2002} that
\begin{align}
[T(p), I(p)]=[L_T(p), L_I(p)]R(p),
\end{align}
where $L_T(p)$ and $L_I(p)$ are the intensities of  ambient light when the images $T$ and $I$ are captured. $R(p)$ is  the ambient reflectance coefficient of surface, and it highly depends on the smoothness or texture of the surface.

Since both the $L_T(p)$ and $L_I(p)$ are  constant in a small neighborhood, $\phi_{\rho}(I)(p)$ and $\phi_{\rho}(T)(p)$ are usually determined by the reflection $R(p)$. Thus, it can be easily derived that
 \begin{align}
 \phi_{\rho}(I)(p)\approx \phi_{\rho}(T)(p) \approx 0.
 \end{align}

 \begin{figure}[!htb]
\centering{
{\includegraphics[width=3.3in]{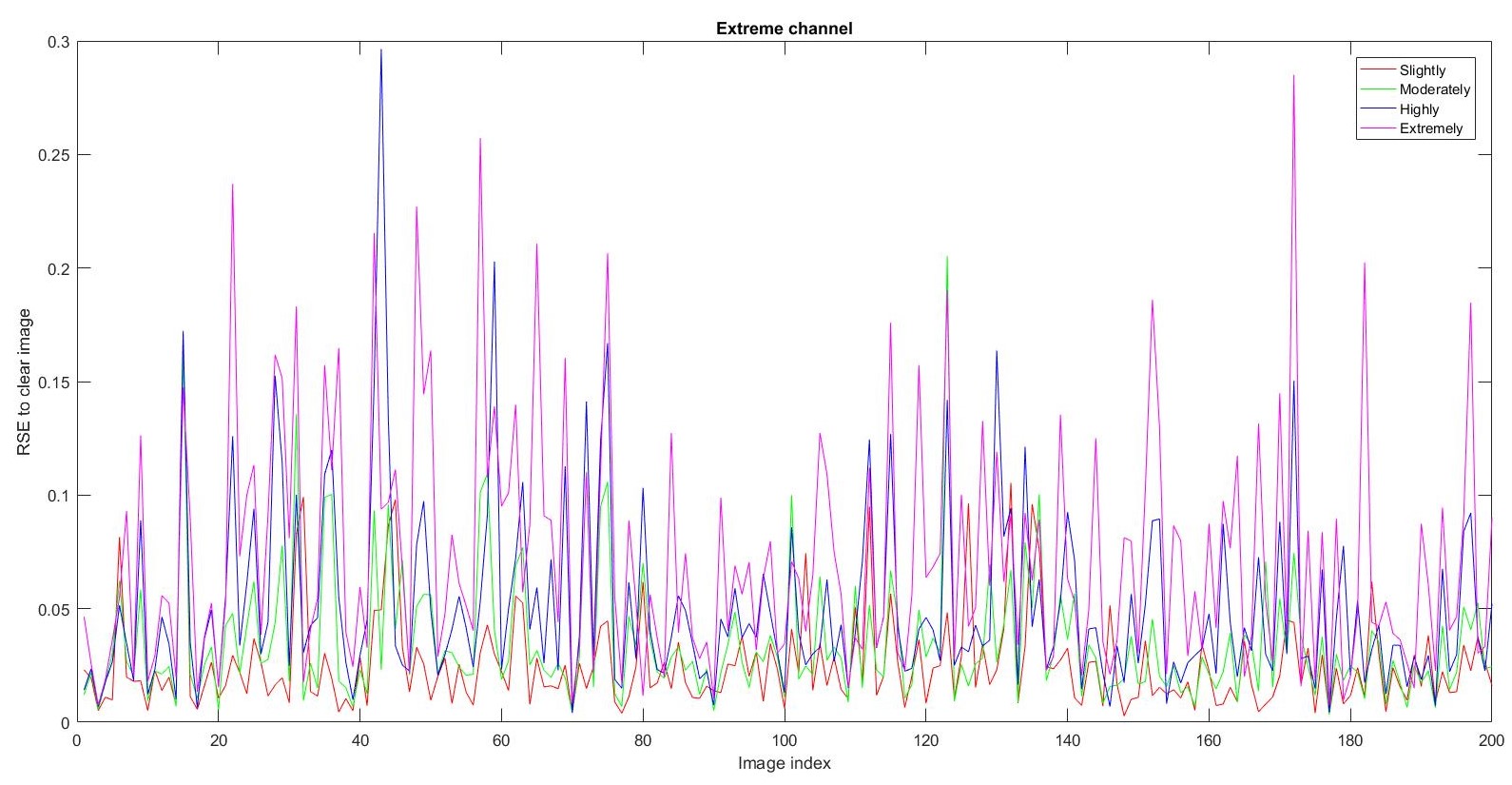}}
}
\caption{MSE between  the extreme channels of hazy and clean images in the dataset \cite{1liuw2021}.}
\label{FigFig3636}
\end{figure}

The mean square error (MSE) between the extreme channels of hazy and clean images in the MRFID dataset \cite{1liuw2021} is shown in Fig. \ref{FigFig3636}. It can be observed that the MSE is usually decreased when the haze degree is reduced from extremely to slightly. The first  loss function is defined by using the extreme channels $\phi_{\rho}(I)$ and $\phi_{\rho}(T)$ as
\begin{equation}
L_e=\frac{1}{WH}\sum_p \|\phi_{\rho}(I)(p)-\phi_{\rho}(T)(p)\|_2^2,
\end{equation}
where $W$ and $H$ are the width and height of the image $I$, respectively. The loss function $L_e$ can guarantee that the haze is well removed in the restored image $I$.

The $L_e$ is different from the corresponding loss function in \cite{1li2020} in the senses that 1) the proposed $L_e$ is on top of the extreme channel (\ref{extremechannel}) while the loss function in \cite{1li2020} is based on the dark channel, and 2) the extreme channel of the restored image is required to match that of the ground-truth image by the $L_e$ while the dark channel of the restored image is required to be zeros in \cite{1li2020}.

The second loss function is defined by using the gradients of the restored image $I$ and the ground-truth image $T$ as
\begin{align}
L_t = \frac{{\displaystyle \sum_{p,c}}(|\nabla_h I_c(p)-\nabla_h T_c(p)|+|\nabla_v I_c(p)-\nabla_v T_c(p)|)}{WH},
\end{align}
where $\nabla_h$ and $\nabla_v$ represent the horizontal and vertical gradients, respectively. The loss function $L_t$ can guarantee that the morphological artifacts are well reduced from the restored image $I$ and the sharpness of the restored image $I$.

The $L_t$ is also different from the corresponding loss function in \cite{1li2020} in the sense that the gradients of the restored image are required to approach those of the ground-truth image $T$ rather than zeros as in \cite{1li2020}. As such, fine details of the restored image $I$ could be preserved better, and the fine details are important for low-level image processing \cite{1kou2016}.

The third loss function is by using the dual-scale dehazing algorithm (\ref{Iij}), and is defined as
\begin{align}
L_r=\frac{\sum_p|I(p)-T(p)|}{WH}+\frac{\sum_p|\{I\}_G^{1}(p)-\{T\}_G^{1}(p)|}{WH}.
\end{align}

All the first three loss functions are combined together as
\begin{equation}
\label{eq14}
L_{cnn}= w_RL_r+w_e L_e +  L_t,
\end{equation}
where $w_R$ and $w_e$ are two constants, and their values are empirically selected as 100 and 100, respectively  if not specified in this paper.

The fourth  adversarial loss function is also dual-scale. One scale is defined by using the restored image $I$ and ground-truth image $T$ as
\begin{align}
L_{adv}^{0}=\log(D_0(T))+\log(1-D_0(I)),
\end{align}
and the other scale is defined via the restored $\{I\}_G^{1}$ and ground-truth $\{T\}_G^{1}$ as
\begin{align}
L_{adv}^{1}=\log(D_1(\{T\}_G^{1}))+\log(1-D_1(\{I\}_G^{1})),
\end{align}
where the PatchGAN in \cite{1isola2017} is adopted to implement the two discriminators $D_0$ and $D_1$.  The overall adversarial loss functions $L_{adv}$ is defined as the sum of $L_{adv}^0$ and $L_{adv}^1$. $L_{adv}$ can guarantee that the textures and reflections of the images $I$ and $T$ are almost the same.

The atmospheric light $A$ and transmission map $t(p)$ are refined by minimizing the following overall loss function:
\begin{align}
\label{wadv}
L_d= w_{adv}L_{adv}+L_{cnn},
\end{align}
where $w_{adv}$ is a constant, and its value is empirically selected as 1.0 if not specified.

Our implementation is on top of a PyTorch framework  with 4 NVIDIA GP100 GPUs. Both mirroring and randomly cropped $128*128$ patches from each input are employed to augment training data. The GANs in Fig. \ref{Fig3} are trained using the proposed loss function (\ref{wadv}) and an Adam optimizer with the batch size as $16$. The learning rate is initially set to $10^{-5}$, then decreased using a cosine annealing schedule. The proposed algorithm is summarized as in the algorithm \ref{algo3}.
\begin{algorithm}
{
\caption{Neural augmented dual-scale single image dehazing}
\label{algo3}
\begin{enumerate}
\item[Step 1.] Estimate the  atmospheric light $A_m$ from the hazy image $Z$ using the hierarchical searching method in \cite{Three}.
\item[Step 2.] Initialize the transmission map $t_0$ by using the DDAP as in the equation (\ref{transmap}).
\item[Step 3.] Reduce the morphological artifacts of $t_0$ via the nonlocal HLA (\ref{myrrmaxij11}), and and generate the initial transmission map $t_m$.
\item[Step 4.] Refine the atmospheric light and transmission map using the GAN in Fig. \ref{Fig3}.
\item[Step 5.] Restore the haze free Lyaplacian pyramid $\{I\}_G^{1}$ and $\{I\}_L^{0}$ via the equation (\ref{Iij}). The pyramid is collapsed to produce the haze-free image $I$.
\end{enumerate}}
\end{algorithm}

It can be easily verified by the images in Fig. \ref{FigFig2} that the dynamic range of the restored image  is higher than that of the hazy image  \cite{1zheng2013,1kouf2018}. Both the global contrast and the local contrast of the restored image are larger than those of the haze image. The restored image usually looks darker than the hazy image, and it can be brightened by using an existing single image brightening algorithm.

\begin{table*}[!htb]
	\centering
	\caption{Average PSNR ($\uparrow$) and SSIM ($\uparrow$) values of 500 outdoor hazy images in the SOTS for different algorithms.}
{\small	\begin{tabular}{c|c|c|c|c|c|c|c|c|c}
		\hline
 &  FD-GAN \cite{1dong2020}&  RefineDNet \cite{1zhao2021} &	FFA-Net \cite{1qin2020} & PSD \cite{1chen2021} & DCP \cite{1he2013}     & HLP \cite{CVPR16} &   MSBDN \cite{1dongh2020}  & DDAP \cite{1LiZG2021}  &    Ours  \\\hline
PSNR	& 20.78 & 20.80 & 32.13 & 15.15  & 17.49  & 18.06  & 30.25	& 17.11  & 21.76\\\hline	
SSIM	& 0.8625 & 0.8981 & 0.9792 & 0.7354  & 0.8555  & 0.8491  & 0.9442	& 0.8405  & 0.9094\\
		\hline
		\end{tabular}}	
    \label{table3}
\end{table*}

\begin{table*}[htb]
	\centering
	\caption{Average DHQI ($\uparrow$) and FADE ($\downarrow$) values of 79 real-world outdoor hazy images for different algorithms.}
{\small	\begin{tabular}{c|c|c|c|c|c|c|c|c|c}
\hline
 &  FD-GAN \cite{1dong2020}& RefineDNet \cite{1zhao2021} &	FFA-Net \cite{1qin2020} & PSD \cite{1chen2021} & DCP \cite{1he2013}     & HLP \cite{CVPR16} &   MSBDN \cite{1dongh2020} & DDAP \cite{1LiZG2021}  &      Ours    \\\hline
 DHQI	&51.00   &57.57  &55.33  &50.60   &51.92   &52.75   &54.32   &60.97  &62.58 \\\hline
 FADE	&0.6261  &0.6690 &1.8289 &0.6679  &0.6771  &0.3980  &1.3745  &0.4600  &0.5883 \\\hline
		\end{tabular}}
    \label{table2}
\end{table*}

\begin{figure*}[!htb]
	\centering
	\includegraphics[width=0.95\textwidth]{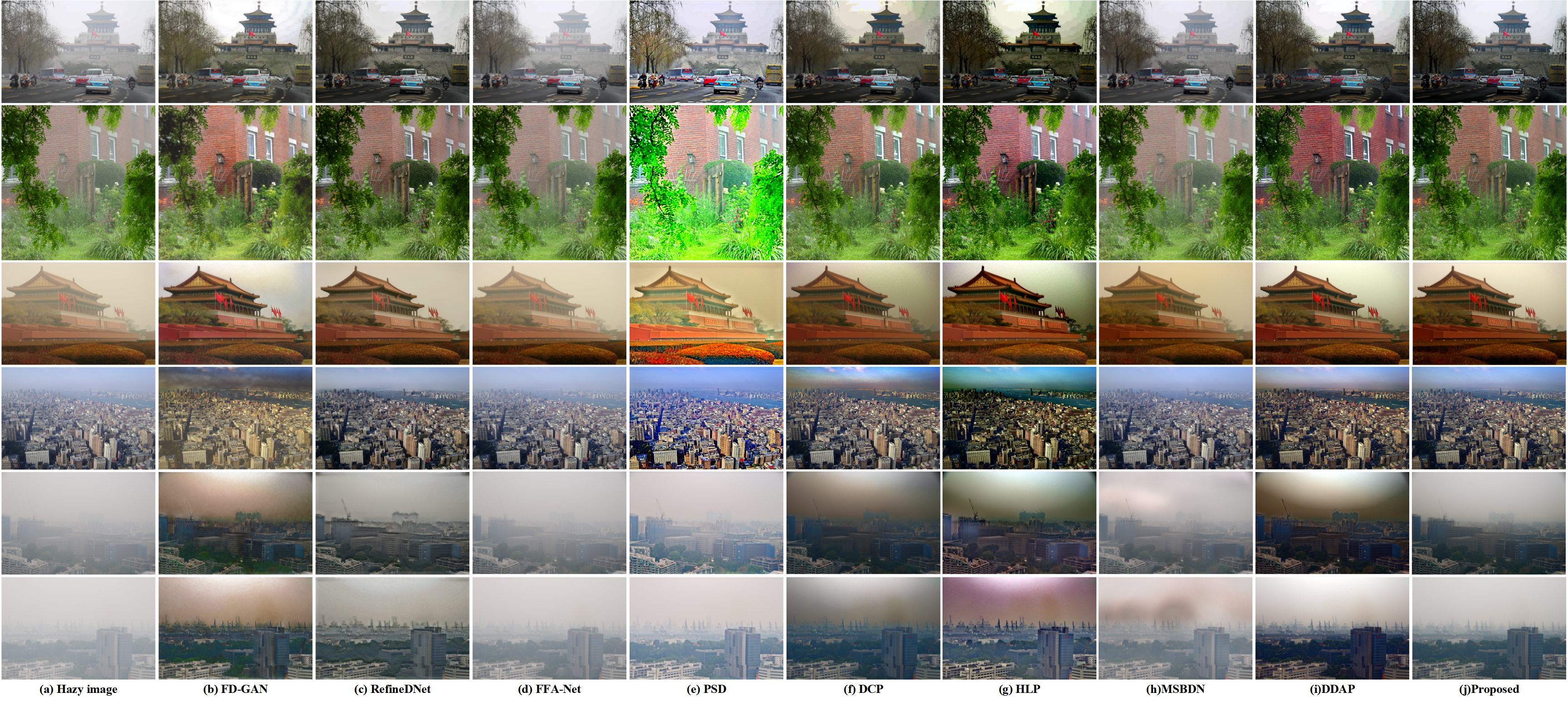}
	\caption{Comparison of different haze removal algorithms. From left to right, hazy images, dehazed images by FD-GAN \cite{1dong2020},
		RefineDNet \cite{1zhao2021}, FFA-Net \cite{1qin2020},  PSD \cite{1chen2021}, DCP \cite{1he2013}, HLP \cite{CVPR16}, MSBDN \cite{1dongh2020}, DDAP \cite{1LiZG2021}, and the proposed one, respectively. }
	\label{Fig6789}
\end{figure*}

\section{Experimental Results}
\label{experiment}
Extensive experimental results on synthetic and real-world hazy images are provided in this section to validate the proposed model-based deep learning framework with emphasis on illustrating how the model-based method and the data-driven one {\it compensate} each other in the proposed neural augmentation.

\subsection{Datasets}

The proposed algorithm is trained by using 500 images with heavy haze that are generated using 500 ground-truth images from the RESIDE datasets \cite{1LiB2018}. Depth is estimated by using the algorithm in \cite{1lizq2018}. The scattering coefficients of 100 images are randomly generated in $[1.2, 2.0]$ and those of the others are randomly selected in $[2.5, 3.0]$. The $A_c$'s are randomly generated in $[0.625, 1.0]$ for the color channel $c$ independently. The hazy images of the 25 outdoor scenes from the MRFID dataset with the corresponding 100 hazy images are selected as the validation set \cite{1Lizg2022}. All these hazy images are randomly selected from the datasets. The proposed framework is trained for 100 epochs and is  tested on the validation dataset after 5 epoches. The test images comprise 500 outdoor hazy images in the synthetic objective testing set (SOTS) \cite{1LiB2018}, and 79 real-world hazy images in \cite{1LiZG2021}  which include 31 images are from the RESIDE datasets \cite{1LiB2018}, as well as 19 images from the reference \cite{1fattal2008} and the Internet.

\begin{figure*}[htb]
	\centering
	\includegraphics[width=0.96\textwidth]{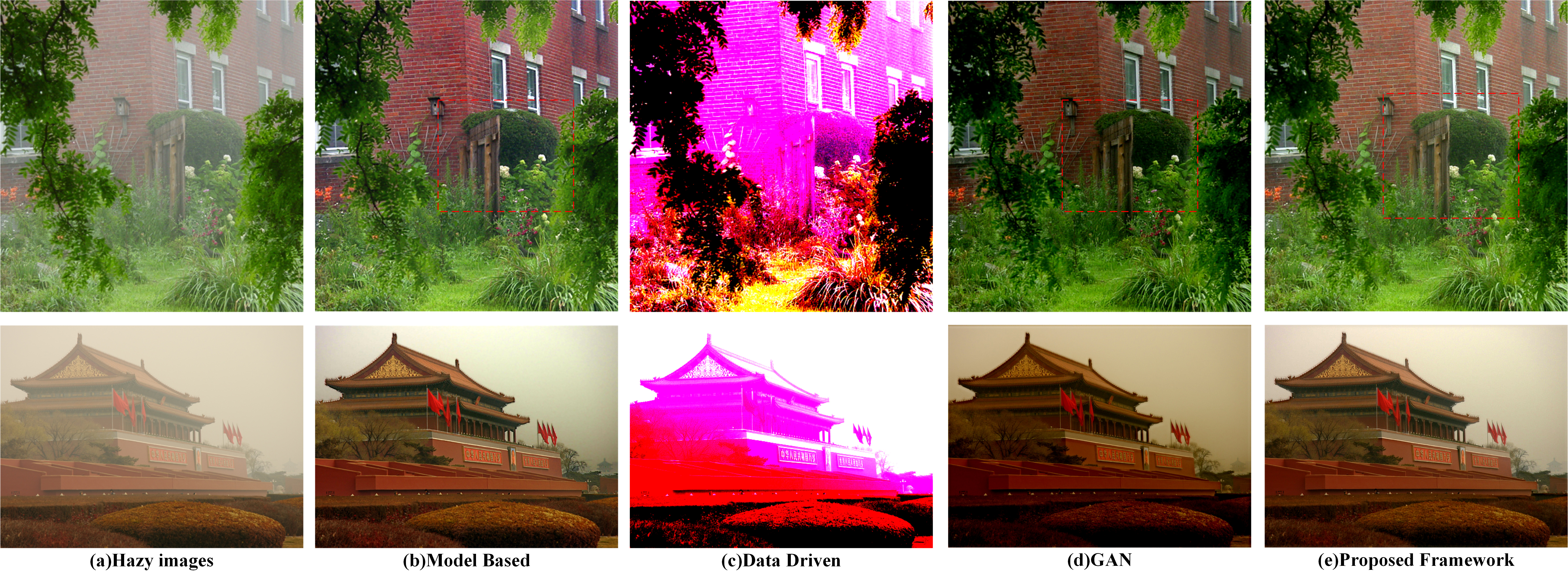}
	\caption{Comparison of different dehazing methods. From left to right, (a) hazy images as well as dehazed images by (b) model-based method, (c) data-driven approach, (d) GAN-based neural augmentation, and (e) the proposed neural augmentation. }
	\label{Fig6}
\end{figure*}

\subsection{Comparison of Different Dehazing Algorithms}

The proposed dehazing algorithm is  compared with eight state-of-the-art dehazing algorithms including RefineDNet \cite{1zhao2021}, FFA-Net \cite{1qin2020}, PSD \cite{1chen2021},  DCP \cite{1he2013}, HLP \cite{CVPR16}, MSBDN \cite{1dongh2020}, FD-GAN \cite{1dong2020}, and DDAP \cite{1LiZG2021}. Among them, the algorithms HLP \cite{CVPR16}, DCP \cite{1he2013}, and DDAP \cite{1LiZG2021} are model-based algorithms, the proposed algorithm and the RefineDNet \cite{1zhao2021} are combination of model-based and data-driven approaches,  while the others are data-driven algorithms. All the results in \cite{1zhao2021,1qin2020,1chen2021,1dongh2020,1dong2020} are generated by their publicly shared codes. The size of the input images  must be dividable by 16 for the MSBDN \cite{1dongh2020}. Therefore, the sizes of the input images are resized for the MSBDN \cite{1dongh2020}, and  the restored images are resized back to the original ones. The sizes of several images exceed the GPU memory when using the FD-GAN \cite{1dong2020}, their sizes are reduced to half of the original ones, and the dehazed images are resized back to their original sizes.

The PSNR and SSIM are first adopted to compare the proposed algorithm with those in \cite{1zhao2021},  \cite{1qin2020}, \cite{1chen2021}, \cite{1he2013}, \cite{CVPR16}, \cite{1LiZG2021}, \cite{1dongh2020},  and \cite{1dong2020} by using 500 synthetic outdoor hazy images in the SOTS  \cite{1LiB2018}. The average PSNR values of the eight algorithms are given in Table \ref{table3}.  The FFA-Net \cite{1qin2020} and  MSBDN \cite{1dongh2020} are two CNN based dehazing algorithms, and they are optimized on top of the $l_1$ and $l_2$ loss functions to provide higher PSNR and SSIM values. However, the dehazed images look a little blurry. This is because the $l_1$ or $l_2$ loss function is minimized by averaging all plausible outputs, which causes blurring. Both the proposed algorithm and the algorithm in \cite{1zhao2021} indeed outperform the model-based algorithms in \cite{CVPR16}, \cite{1LiZG2021}, and \cite{1he2013} from the PSNR and SSIM points of view.

The quality index DHQI in \cite{1min2019} and FADE in \cite{1choi2015} are then adopted to compare the proposed algorithm with those in \cite{1zhao2021},  \cite{1qin2020}, \cite{1chen2021}, \cite{1he2013}, \cite{CVPR16}, \cite{1LiZG2021}, \cite{1dongh2020},  and \cite{1dong2020} by using the 79 real-world hazy images in \cite{1LiZG2021}. The average DHQI and FADE values of the 79 real-world outdoor hazy images are given in Table \ref{table2}. The proposed algorithm is ranked first from the DHQI point of view and third from the FADE point of view. It outperforms all the data-driven algorithms and the algorithm in \cite{1zhao2021} from the DHQI and FADE points of view.

Finally, all these dehazing algorithms are compared subjectively as in Fig. \ref{Fig6789}. Readers are invited to view to electronic version of figures and zoom in them so as to better appreciate differences among all images.  Although the FFA-Net \cite{1qin2020} and  MSBDN \cite{1dongh2020} achieve higher PSNR and SSIM values, their dehazed results are a little blurry and they are not photo-realistic. In addition, the haze is not reduced well if it is heavy. The DCP  \cite{1he2013}, HLP \cite{CVPR16}, RefineDNet \cite{1zhao2021}, FD-GAN \cite{1dong2020}, DDAP \cite{1LiZG2021}, and the proposed algorithm can be applied to generate  photo-realistic images.  There are visible morphological artifacts in the restored images by the PSD \cite{1chen2021}, RefineDNet \cite{1zhao2021} and FD-GAN \cite{1dong2020}. Textures generated by the RefineDNet \cite{1zhao2021} and FD-GAN \cite{1dong2020} are different from the real ones. The DCP  \cite{1he2013}, HLP \cite{CVPR16}, and DDAP \cite{1LiZG2021} restore sharper images at the expense of color distortion, and noise is amplified in sky regions by the DCP  \cite{1he2013} and HLP \cite{CVPR16}. All these problems are overcome by the proposed algorithm. However, the proposed algorithm has a limitation to restore those far-away objects in some images with heavy haze.

\begin{table*}[htb]
	\begin{center}
		\centering
		\caption{Ablation study on different key components of the proposed algorithm ($\uparrow$: larger is better, $\downarrow$: smaller is better)}
		{\small \tabcolsep12pt\begin{tabular}{c|cccc|cccc}\hline
			Case & Model-based &  Dual-scale & $L_{cnn}$ & Refining $A$   &   SSIM ($\uparrow$)    &  PSNR ($\uparrow$) &  DHQI ($\uparrow$) &  FADE ($\downarrow$)\\\hline
			1 & Y & N & N  & N   & 0.841  & 17.11       & 60.97     &0.4600\\
			2 & N & Y & Y & Y    & 0.385  & 8.09        & 34.46     &0.5071 \\
			3 & Y & N & Y & Y    & 0.8734  & 19.83       & 61.28     &0.6328\\
			4 & Y & Y & N & Y    & 0.8935  & 21.03       & 60.62     &0.6716\\
			5 & Y & Y & Y & N    & 0.8665  & 20.64       & 60.58     &0.6062 \\
			6 & Y & Y & Y  & Y   & 0.9094  & 21.76       & 62.58     &0.5883\\\hline
		\end{tabular}}
		\label{tab3}
	\end{center}
\end{table*}

\subsection{Ablation Studies}

\begin{figure}[!htb]
	\centering{
		{\includegraphics[width=3.3in]{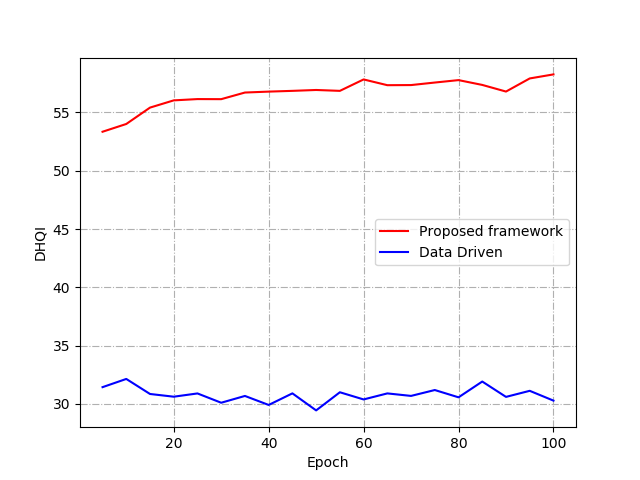}}
	}
	\caption{Comparison of convergence speed for the data-driven approach and the proposed neural augmentation. The data-driven approach using the GAN in  Fig. \ref{Fig3} does not converge.}
	\label{FigFig13636}
\end{figure}

{\it 1) Comparison between the data-driven approach (Case 2 in Table \ref{tab3}) and the proposed neural augmentation (Case 6 in Table \ref{tab3})}: In order to verify the contrition of the estimated $A$ and $t$ from the model-based method in the proposed neural augmentation dehazing algorithm,  the $A$ and $t$ are learnt directly by the same GAN in  Fig. \ref{Fig3} without utilizing the estimated values from the model-driven method. As shown in Fig. \ref{FigFig13636}, the proposed framework indeed converges very fast while the data-driven approach (i.e. g the GAN in  Fig. \ref{Fig3}) does not converge. It was also pointed out in \cite{1gan2014,1liu2017,1heusel2017,1naga2017} that the stability of the GAN  could be an issue. Therefore,  the stability of the proposed neural augmentation is improved by the accurate model-based components $A_m$ and $t_m(p)$. This is also illustrated in the appendix.

{\it 2) Comparison between single-scale and dual-scale dehazing algorithms}: Besides the dual-scale dehazing algorithm, another one is a single-scale dehazing algorithm (Case 3 in Table \ref{tab3}).  As demonstrated in Table  \ref{tab3}, the proposed dual-scale dehazing algorithm can indeed improve the proposed dehazing algorithm.

{\it 3) Comparison between two different training methods}: Besides the proposed training method, one alternative training method is  to use the loss function $L_{adv}$ only, and this results in a GAN-based refinement (Case 4 in Table \ref{tab3}).  Although the GAN-based refinement can generate sharp images, the color of the restored images could be over-saturated as shown in Fig. \ref{Fig6}. The color distortion is reduced by the proposed training method.

{\it 4) Comparison between two different refinement methods}:  Inaccurate estimations of transmission map $t$ could result in morphological artifacts and color distortion in the restored images as shown in Fig. \ref{FigFig2}. Thus, the transmission map $t$ is always refined. There are two alternative strategies,  refine the atmospheric light $A$ and do not refine it (Case 5 in Table \ref{tab3}). The latter is widely chosen in the model-based algorithms \cite{1he2013,CVPR16,1LiZG2021} and the model-based deep learning one in \cite{1zhao2021}. As shown in Fig. \ref{FigFig3636} (c), the refinement of the atmospheric light $A$ can alleviate possible color and brightness distortions.

  \begin{figure}[!htb]
 	\centering{
 		{\includegraphics[width=3.3in]{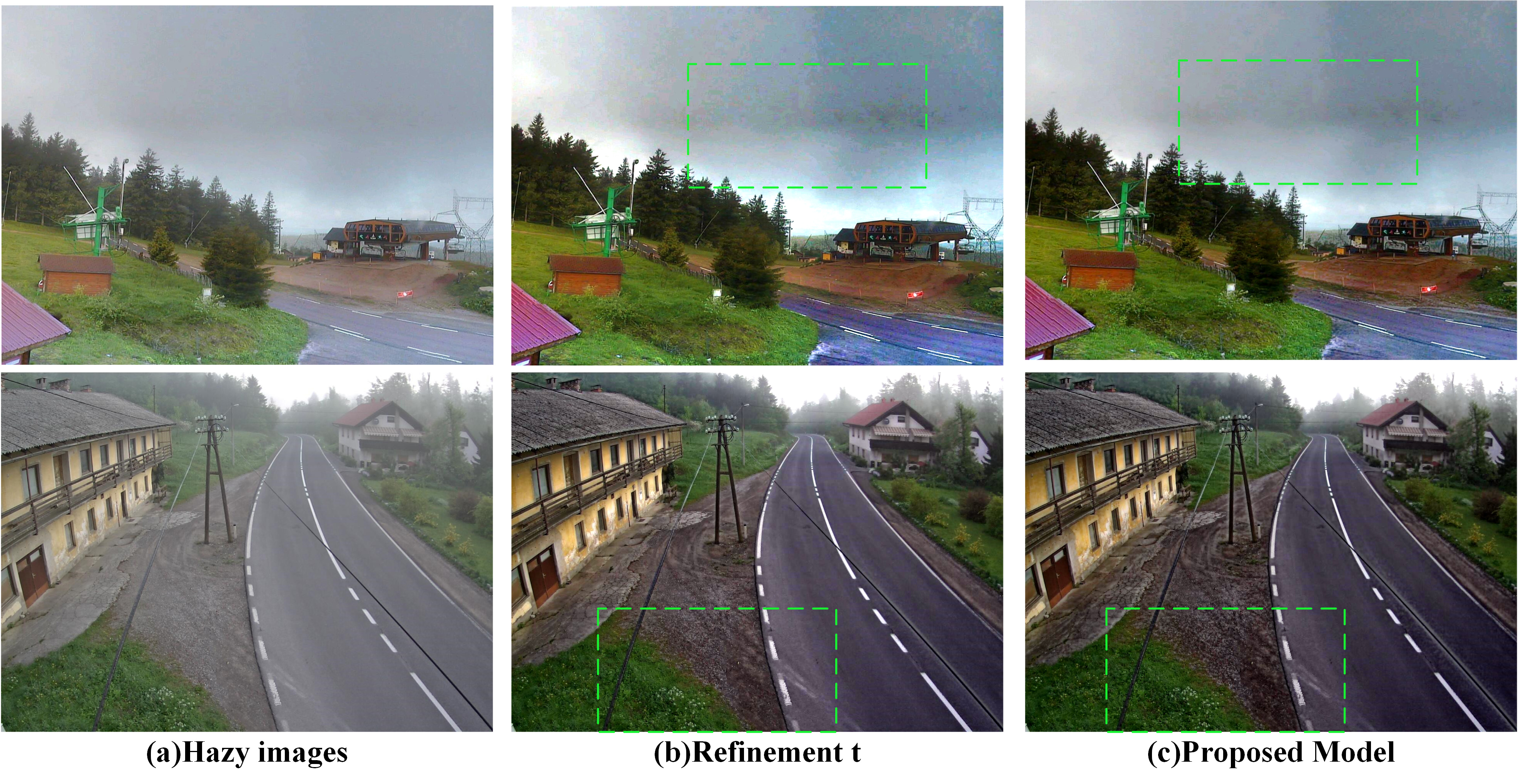}}
 	}
 	\caption{Comparison between the proposed framework with another framework which only refines $t$. (a) hazy image as well as dehazed images by (b) the framework which only refines $t$, and (c) the proposed framework.}
 	\label{FigFig3636}
 \end{figure}

Overall, it can be shown from experimental results  in Table \ref{tab3} that the proposed components are useful to improve the performance of the proposed neural augmentation based dehazing algorithm.

\section{Conclusion Remarks and Discussion}
\label{conclusion}

In this paper, a new type of single image dehazing algorithm is introduced by using model-based deep learning frameworks. Both transmission map and atmospheric light are obtained by a neural augmentation  which consists of model-based initialization and data-driven refinement. They are then applied to restore a haze-free image. Experimental results validate that the proposed algorithm removes haze well from  the synthetic and real-world hazy images. The proposed neural augmentation reduces the number of training data significantly, and the proposed neural augmentation framework converges faster than the corresponding data-driven approach. It is thus more friendly to domain adaptation and continual learning. It is worth noting that this paper focused on day-time hazy images. The proposed framework will be extended to study night-time hazy images \cite{1yangm2018} in our future research.

\section*{Appendix: Convergence Analysis of the Proposed Neural Augmentation Framework}
Control theory in \cite{1lizg2001,1khalil2002} is borrowed for the convergence analysis of the proposed neural augmentation framework. Let $\theta$ denote the vector of all network parameters, $\theta_{\tau}$ be the time-dependence of the parameters, and $\eta_{\tau}$ be the learning rate.

It can be derived from the equations (\ref{eq1}) and (\ref{softmax}) that
\begin{align}
\label{neuralaugmentation123}
I_{\tau}(p)=I_m(p)+I_{d,\tau}(p),
\end{align}
where $I_m(p)$ and $I_{d,\tau}(p)$ are computed as
\begin{align*}
\left\{\begin{array}{l}
I_m(p)=\frac{Z(p)-A_m}{t_m(p)}+A_m\\
I_{d,\tau}(p)=\frac{Z(p)-(A_m+A_{d,\tau})}{t_m(p)+t_{d,\tau}(p)}+A_m+A_{d,\tau}-I_m(p)
\end{array}
\right..
\end{align*}

For brevity of notation, $I_{\tau}$ also denotes the vector of the restored image.   It can be derived that \cite{1Lee2019}
\begin{align}
\left\{\begin{array}{l}
\dot{\theta}_{\tau}=-\eta_{\tau}(\nabla_{\theta}I_{\tau})^T\nabla_{I_{\tau}}L_d\\
\dot{I}_{\tau}=\nabla_{\theta}I_{\tau}\dot{\theta}_{\tau}=-\eta_{\tau} \Theta_{\tau}\nabla_{I_{\tau}}L_d
\end{array}
\right.,
\end{align}
where the matrix $\Theta_{\tau}$ is   $\nabla_{\theta}I_{\tau}(\nabla_{\theta}I_{\tau})^T$.

Similar to \cite{1Lee2019}, consider the case that the loss function $L_d$ is defined by the $l_2$ norm and the learning rate $\eta$ is fixed. It follows that
\begin{align}
\dot{I}_{\tau}=-\eta \Theta_{\tau}(I_{\tau}-T).
\end{align}
{\bf If the models $t_{m}$ and $A_m$ in the equation (\ref{softmax}) are accurate}, {\it $I_m$ in the equation (\ref{neuralaugmentation123}) then corresponds to the data-driven approach $I_{d,\tau}$ with an initial $\theta_0$ and the $\theta_0$  is in a neighborhood of the optimal $\theta^*$}. By using the first-order Taylor expansion, the neural augmentation framework (\ref{neuralaugmentation123}) is approximated by
\begin{align}
I_{\tau}= I_m+\nabla_{\theta} I_{d,\tau}|_{\theta=\theta_0}(\theta-\theta_0).
\end{align}
Let the matrix $\nabla_{\theta}I_{d,\tau}|_{\theta=\theta_0}(\nabla_{\theta}I_{d,\tau}|_{\theta=\theta_0})^T$ be denoted as  $\Theta_0$ which is  symmetric positive semidefinite. Since the matrix $\Theta_0$ is constant throughout training, it can be derived that
\begin{align}
\nonumber
I_{d,\tau}=\exp(-\eta \Theta_0 \tau)(I_m-T)+T-I_m.
\end{align}
Since $\|I_m-T\|$ is smaller than $\|T\|$, the proposed neural augmentation framework converges faster than the corresponding data-driven approach. It is worth noting that the analysis will be more complicated if the loss function is not the $l_2$ loss.

It was pointed out in \cite{1gan2014,1liu2017,1heusel2017,1naga2017} that the GAN is usually locally stable. Therefore, the accuracy of the model-based components $A_m$ and $t_m(p)$ in the equation (\ref{softmax}) is also important for the stability of the proposed neural augmentation. The proposed neural augmentation framework is actually a switched system \cite{1lizg2001} because the learning rate $\eta_{\tau}$ is  initially set to $10^{-5}$, and then decreased using a cosine annealing schedule.


\begin{thebibliography}{1}

\bibitem{1nara2003}
S. G. Narasimhan and S. K. Nayar,
\newblock ``Contrast restoration of weather degraded images,"
\newblock IEEE Trans. On Pattern Analysis and Machine Learning, vol. 25, no. 6, pp. 713-724, Jun. 2003.

\bibitem{1garg2004} K. Garg and S. K. Nayar,
\newblock ``Detection and removal of rain from videos,"
\newblock in {\it  IEEE/CVF  Conference on Computer Vision and Pattern Recognition}, 2004.


\bibitem{1ancuti2013}
C. O.  Ancuti and C. Ancuti,
\newblock ``Single image dehazing by multi-scale fusion,"
\newblock IEEE Trans. on Image Processing, vol 22, no. 8, pp. 3271--3282, Aug. 2013.

\bibitem{He}
K. He, J. Sun, and X. Tang,
\newblock ``Single image haze removal using dark channel prior,"
\newblock IEEE Trans. On Pattern Analysis and Machine Learning, vol 33, no. 12, pp. 2341-2353, Dec. 2011.



\bibitem{1zheng2013}
J. Zheng, Z. Li, Z. Zhu, S. Wu, and S. Rahardja,
\newblock ``Hybrid patching for a sequence of differently exposed images with moving objects,"
\newblock IEEE Trans. on Image Processing, vol. 22, no. 12, pp. 5190-5201, Dec. 2013.

\bibitem{1kouf2018} F. Kou, Z. Wei, W. Chen, X.  Wu, C. Wen, and Z. Li,
\newblock ``Intelligent detail enhancement for exposure fusion,"
\newblock IEEE Trans. on Multimedia, vol. 20, no. 2, pp. 484--495, Feb. 2018.

\bibitem{1fattal2008} R. Fattal,
\newblock ``Dehazing using color-lines,"
\newblock ACM Trans. on Graphics, vol 34, no. 1, articale 13, Jan. 2014.

\bibitem{1zhuq2015} Q. Zhu, J. Mai, and L. Shao,
\newblock ``A fast single image haze removal algorithm using color attenuation prior,"
\newblock  IEEE Trans. on Image Processing, vol. 24, no. 11, pp. 3522-3533,  Nov. 2015.

\bibitem{CVPR16}
D.  Berman, T.  Treibitz, and S. Avidan,
\newblock ``Non-local Image Dehazing,"
\newblock in {\it IEEE/CVF Conference on Computer Vision and Pattern Recognition (CVPR)},  pp. 1674-1682, 2016.


\bibitem{1LiZG2021}	Z. Li, H.  Shu, and C. Zheng,
 \newblock ``Multi-scale single image dehazing Using Laplacian and Gaussian pyramids,"
 \newblock IEEE Trans. on Image Processing, vol. 30, no. 12, pp. 9270-9279, Dec. 2021.


\bibitem{Ren}
W.  Ren, S.  Liu, H.  Zhang, J.  Pan, X. Cao, and M. Yang,
\newblock ``Single image dehazing via multi-scale convolutional neural networks,"
\newblock in {\it European Conference on Computer Vision}, pp 154-169, Sept. 2016.


\bibitem{cai}
B.  Cai, X.  Xu, K. Jia, C. Qing, and D.  Tao,
\newblock ``DehazeNet: an end-to-end system for single image haze removal,"
\newblock IEEE Trans. on Image Processing, vol. 25, no. 11, pp. 5187-5198, Nov. 2016.


\bibitem{ICCV17}
B. Li, X. Peng, Z. Wang, J.  Xu, and D. Feng,
\newblock ``AOD-Net: all-in-one dehazing network,"
\newblock in {\it IEEE/CVF International Conference on Computer Vision}, pp. 4780-4788, Oct. 2017.

\bibitem{1qin2020} X.  Qin, Z. Wang, Y.  Bai, X. Xie, and H. Jia,
\newblock ``FFA-Net: Feature fusion attention network for single image dehazing,"
\newblock in {\it AAAI Conference on Artificial Intelligence}, pp. 11908-11915, 2020.

\bibitem{1dongh2020} H. Dong, J.  Pan, L.  Xiang, Z.  Hu, X. Zhang, F.  Wang, and M. Yang,
\newblock ``Multi-scale boosted dehazing network with dense feature fusion,"
\newblock in {\it IEEE/CVF Conference on Computer Vision and Pattern Recognition(CVPR)}, 2020.

\bibitem{Qu_2019_CVPR}
Y.  Qu, Y. Chen, J.  Huang, and Y. Xie,
\newblock ``Enhanced Pix2pix dehazing network,"
\newblock in {\it IEEE/CVF Conference on Computer Vision and Pattern Recognition (CVPR)}, pp. 8152-8160, Jun. 2019.



\bibitem{koschmider} H. Koschmieder,
\newblock ``Theorie der horizontalen sichtweite,"
\newblock  in {\it Proc. Beitrage Phys. Freien Atmos.},  pp. 33-53, 1924.

\bibitem{glots2020}
A.   Golts, D. Freedman, and M. Elad,
\newblock ``Unsupervised single image dehazing using dark channel prior loss,"
\newblock IEEE Trans. on Image Processing, 29, pp. 2692 - 2701, 2019.

\bibitem{1liuw2021} W. Liu, F. Zhou, T. Lu, J. Duan, and G. Qiu,
\newblock ``Image defogging quality assessment: real-world database and method,"
\newblock IEEE Trans. on Image Processing, 30, pp. 1762--190, 2021.

\bibitem{1chen2021} Z. Chen, Y.  Wang, Y. Yang, and D. Liu,
\newblock ``PSD: principled synthetic-to-real dehazing guided by physical priors,"
\newblock in {\it IEEE/CVF Conference on Computer Vision and Pattern Recognition}, 2021.

\bibitem{Three}
J. H. Kim, W. D. Jang, J. Y. Sim, and C. S. Kim,
\newblock ``Optimized contrast enhancement for real-time image and video dehazing,"
\newblock Journal of Visual Communication and Image Representation, vol. 24, no. 3, pp. 410-425, Apr. 2013.

\bibitem{1hornik1989} K. Hornik, M.  Stinchcombe, H. White, et al.,
\newblock ``Multilayer feedforward networks are universal approximators,"
\newblock Neural Networks, vol. 2, no. 5, pp. 359-366, May 1989.

\bibitem{1nir2021} N. Shlezinger, J. Whang, Y. C. Eldar, and A.  G. Dimakis,
\newblock ``Model-based deep learning,"
\newblock arXiv: 2012.08405v2 [eess.SP] 27 Jun 2021.

\bibitem{1gan2014}
I. J. Goodfellow, J. Pouget-Abadie, M. Mirza, B. Xu, D. Warde-Farley, S. Ozair, A. Courville, and Y. Bengi,
\newblock ``Generative adversarial nets,"
\newblock  in {\it NIPS 2014}, Canada, Dec. 2014.

\bibitem{1Lizg2022} Z. Li, C. Zheng, H. Shu, and S. Wu,
\newblock ``Model-based single image deep dehazing,"
\newblock in {\it IEEE International Conference on Image Processing}, Oct. 2022.

\bibitem{CycleISP} S. Zamir, A. Arora, S. Khan, M.  Hayat, F. Khan, M. Yang, and L. Shao,
\newblock ``CycleISP: real image restoration via improved data synthesis,"
\newblock in {\it IEEE/CVF Conference on Computer Vision and Pattern Recognition}, 2020.

\bibitem{1raha2018} N. Rahaman, A. Baratin, D. Arpit, F. Draxler, M. Lin, F. Hamprecht, Y. Bengio, and A. Courville,
\newblock ``On the spectral bias of neural networks,"
\newblock in {\it  Proceedings of the 36th International Conference on Machine Learning}, pp. 5301-5310, 2019.


\bibitem{1isola2017} P. Isola, J. Zhu, T. Zhou, and A. A. Efros,
\newblock ``Image-to-image translation with conditional adversarial networks,"
\newblock in {\it IEEE/CVF Conference on Computer Vision and Pattern Recognition}, pp. 1125-1134, 2017.

\bibitem{1LiB2018} B. Li, W. Ren, D. Fu, D. Tao,  D. Feng, W. Zeng, and Z. Wang,
\newblock ``Benchmarking single-image dehazing and beyond,"
\newblock IEEE Trans. on Image Processing vol. 28, no. 1, pp. 492–505, Jan. 2018.

\bibitem{1lizq2018} Z. Li and N. Snavely,
\newblock ``Megadepth: Learning single-view depth prediction from internet photos,"
\newblock in  {\it  IEEE/VF Conference on
Computer Vision and Pattern Recognition}, pp. 2041–2050, 2018.

\bibitem{2zheng2020} C. B. Zheng, Z. G. Li, Y. Yang, and S. Q. Wu,
 \newblock ``Exposure interpolation via hybrid learning,"
 \newblock in {\it International Conference on Acoustics, Speech, and Signal Processing}, pp. 2098-2102, 2020.

\bibitem{1zheng2020} C. Zheng, Z. Li, Y. Yang, and S. Wu,
\newblock ``Single image brightening via multi-scale exposure fusion with hybrid learning,"
\newblock IEEE Trans. on Circuits and Systems for Video Technology, vol. 31, no. 4, pp. 1425-1435, Apr. 2021.

\bibitem{1zhao2021} S. Zhao, L. Zhang, Y. Shen, and Y. Zhou,
\newblock ``RefineDNet: A weakly supervised refinement framework for single image dehazing,"
\newblock IEEE Trans. on Image Processing, vol. 30, pp. 3391-3404, 2021.

\bibitem{1dong2020} Y. Dong, Y. Liu, H. Zhang, S. Chen, and Y. Qiao,
\newblock ``FD-GAN: Generative adversarial networks with fusiondiscriminator for single image dehazing,"
\newblock in {\it Proceedings of the AAAI Conference on Artificial Intelligence}, pp. 10728-10736, 2020.

\bibitem{1min2019}
X.  Min, G. Zhai, K. Gu, X. Yang, and X. Guan,
\newblock ``Objective quality evaluation of dehazed images,"
\newblock  IEEE Trans. on Intelligent Transportation Systems, vol. 20, no. 8, pp. 2879-2892, Aug. 2019.

\bibitem{1choi2015} L. Kwon Choi, J. You, and A. C. Bovik,
\newblock ``Referenceless prediction of perceptual fog density and perceptual image defogging,"
\newblock IEEE Trans. on Image Processing, vol. 24, no. 11, pp. 3888-3901, Nov. 2015.

\bibitem{1lizg2001}	Z.  Li, Y. C. Soh, and C.  Wen,
 \newblock ``Robust stability of a class of hybrid nonlinear systems,"
 \newblock IEEE Trans. on Automatic Control, vol. 46, no. 6, pp.897-903, Jun. 2001.

\bibitem{1khalil2002} H. K. Khalil and J. Grizzle,
\newblock Nonlinear system, Prentice Hall, 2002.

\bibitem{Li2015}
Z. Li and J. Zheng,
\newblock ``Single image de-Hazing using globally guided image filtering,"
\newblock IEEE Trans. on Image Processing, vol. 27, no. 1, pp. 442-450, Jan. 2018.

\bibitem{1he2013} K. He, J. Sun, and X. Tang,
\newblock ``Guided image filtering,"
\newblock  IEEE Trans. on Pattern Analyis and Machine Intelligence, vol. 35, no. 6, pp. 1397-1409, Jun. 2013.

\bibitem{Li2014}
Z.  Li, J. Zheng, Z. Zhu, W. Yao, and S. Wu,
\newblock ``Weighted guided image filtering,"
\newblock IEEE Trans. on Image Processing, vol. 24, no. 1, pp. 120-129, Jan. 2015.


\bibitem{1orchard1991}
 M. T. Orchard and C. A. Bouman,
 \newblock ``Color quantization of images,"
  \newblock IEEE Trans. on Signal Processing, vol. 39, no. 12, pp. 2677--2690, Dec. 1991.

\bibitem{1farb2008} Z. Farbman, R.  Fattal, D. Lischinski, and R. Szeliski,
\newblock ``Edge-preserving decompositions for multi-scale tone and detail manipulation,"
\newblock ACM Trans. on Graphics, vol. 27, no. 3, pp. 249-256, Jul. 2008.

\bibitem{1zhang2018} H. Zhang and V. M. Patel,
\newblock ``Densely connected pyramid dehazing network,"
\newblock in {\it IEEE/CVF Conference on Computer Vision and Pattern Recognition}, pp. 3194-3203, 2018.

\bibitem{1shao2020}
Y.  Shao, L. Li, W. Ren, C. Gao, and N. Sang,
\newblock ``Domain adaptation for image dehazing,"
\newblock in {\it IEEE/CVF Conference on Computer Vision and Pattern Recognition}, pp. 2808-2817, 2020.

\bibitem{1yang2018} X. Yang, Z. Xu, and J. Luo,
\newblock ``Towards perceptual image dehazing by physics-based disentanglement and adversarial training,"
\newblock in {\it Proceedings of the AAAI Conference on Artificial Intelligence}, 2018.


\bibitem{1li2020}
L. Li, Y. Dong, W. Ren, J. Pan, C. Gao, N. Sang, and M. Yang,
\newblock ``Semi-supervised image dehazing,"
\newblock IEEE Trans. on Image Processing, vol. 29, pp. 2766-2779, 2020.

\bibitem{1wang2020} W. Wang, Z. G. Li, S. Q. Wu, and L. C. Zeng,
\newblock ``Haze image decolorization with color contrast restoration,"
\newblock IEEE Trans. on Image Processing, vol. 29, no. 1, pp. 1776-1787, Jan. 2020.

\bibitem{1liu2017}
S.  Liu, O.  Bousquet, and K. Chaudhuri,
\newblock ``Approximation and convergence properties of generative
adversarial learning,"
\newblock in {\it Advances in Neural Information Processing Systems 31 (NIPS 2017)},
2017. arXiv:1705.08991.

\bibitem{1heusel2017} M. Heusel, H. Ramsauer, T.  Unterthiner, B.  Nessler, and S. Hochreiter,
\newblock ``GANs trained by a two time-scale update rule converge to a local Nash equilibrium,"
\newblock in {\it NIPS'17: Proceedings of the 31st International Conference on Neural Information Processing SystemsDecember 2017}, pp. 6629-6640.

\bibitem{1naga2017}
V.  Nagarajan and J. Z.Kolter,
\newblock ``Gradient descent GAN optimization is locally stable,"
\newblock  in {\it Advances in Neural Information Processing Systems 31
(NIPS 2017)}. arXiv e-prints, arXiv:1706.04156, 2017.


\bibitem{1gonz2002}  R.  C. Gonzalez and R. E. Woods,
\newblock  ``Digital image processing,"
\newblock Englewood Cliffs, NJ, USA: Prentice-Hall, 2002.

\bibitem{1kou2016} F. Kou, W. H. Chen, Z. G. Li, and C. Y. Wen, 
\newblock ``Content adaptive image detail enhancement,"
\newblock  IEEE Signal Processing Letters, vol. 22,  no. 2,  pp. 211-215,  Feb. 2015.

\bibitem{1bell2015}
S. Bell, P. Upchurch, N. Snavely, and K. Bala,
\newblock ``Material recognition in the wild with the materials in context database,"
\newblock in {\it IEEE/CVF Conference on Computer Vision and Pattern Recognition}, 2015.

\bibitem{1burt1983} P. J. Burt and E. H. Adelson,
\newblock ``The Laplacian pyramid as a compact image code,"
\newblock IEEE Trans. Communication, vol. 31, no. 4, pp. 532-540,
Apr. 1983.

\bibitem{1yangm2018} 	M. Yang, J. Liu, and Z. Li,
 \newblock ``Super-pixel based single nighttime image haze removal,"
 \newblock IEEE Trans. on Multimedia, vol. 20, no. 11, pp. 3008-3018, Nov. 2018.

 \bibitem{1Lee2019} J.  Lee, L. Xiao, S. Schoenholz, Y. Bahri, R. Novak, J. SohlDickstein, and J. Pennington,
\newblock ``Wide neural networks of any depth evolve as linear models under gradient descent,"
\newblock in {\it NeurIPS}, 2019.


\end{thebibliography}
\end{document}